\documentclass{article}

\usepackage[utf8]{inputenc}
\usepackage[T1]{fontenc}
\usepackage[table]{xcolor}
\usepackage{iclr2026_conference,times}
\usepackage{hyperref}
\usepackage{url}
\usepackage{booktabs}
\usepackage{amsmath,amssymb}
\usepackage{array}
\usepackage{microtype}
\iclrfinalcopy

\title{TimeSeek: Temporal Reliability of Agentic Forecasters}

\author{
\makebox[\textwidth][c]{%
\begin{tabular}{@{}c@{\hspace{1.2em}}c@{\hspace{1.2em}}c@{}}
\begin{tabular}[t]{c}
Dennis Lee\thanks{Equal contribution.}\\
{\normalfont\small Automorphic Labs}\\
{\normalfont\small \mbox{}}\\
{\normalfont\scriptsize \nolinkurl{432dennis@gmail.com}}
\end{tabular}
&
\begin{tabular}[t]{c}
Hamza Mostafa\footnotemark[1]\\
{\normalfont\small Cheriton School of Computer Science}\\
{\normalfont\small University of Waterloo}\\
{\normalfont\scriptsize \nolinkurl{hamza.mostafa@uwaterloo.ca}}
\end{tabular}
&
\begin{tabular}[t]{c}
Om Shastri\footnotemark[1]\\
{\normalfont\small The Wharton School}\\
{\normalfont\small University of Pennsylvania}\\
{\normalfont\scriptsize \nolinkurl{omshastri@gmail.com}}
\end{tabular}
\end{tabular}}
}

\newcommand{\pos}[1]{\textbf{#1}}
\newcommand{\negcell}[1]{\cellcolor{gray!20}#1}

\begin{document}

\maketitle

\begin{abstract}
We introduce TimeSeek, a benchmark for studying how the reliability of agentic LLM forecasters changes over a prediction market's lifecycle. We evaluate 10 frontier models on 150 CFTC-regulated Kalshi binary markets at five temporal checkpoints, with and without web search, for 15{,}000 forecasts total. Models are most competitive early in a market's life and on high-uncertainty markets, but much less competitive near resolution and on strong-consensus markets. Web search improves pooled Brier Skill Score (BSS) for every model overall, yet hurts in 12\% of model-checkpoint pairs, indicating that retrieval is helpful on average but not uniformly so. Simple two-model ensembles reduce error without surpassing the market overall. These descriptive results motivate time-aware evaluation and selective-deference policies rather than a single market snapshot or a uniform tool-use setting.
\end{abstract}

\section{Introduction}

When should an autonomous AI agent's forecast be trusted relative to a prediction market? Existing LLM forecasting benchmarks usually evaluate a single snapshot per question or market \citep{Nel2025,KargerEtAl2024,YeEtAl2024}. That setup obscures an important temporal issue: prediction markets aggregate information continuously, so early prices can reflect sparse participation while late prices often incorporate much richer information. A model that is competitive early but weak late has a very different reliability profile from one that underperforms uniformly.

We study this question with TimeSeek, a benchmark built from 150 CFTC-regulated Kalshi binary markets evaluated at five lifecycle checkpoints under two conditions: with and without web search. This yields 15{,}000 model forecasts across 10 frontier models. Rather than treating ``LLMs vs.\ markets'' as a single aggregate comparison, our goal is to characterize when models are relatively competitive, when retrieval helps, and when deferring to the market remains the stronger default. Throughout, we treat these comparisons as descriptive: the paper reports point estimates, while uncertainty quantification for subgroup analyses is left to future work. Our main contributions are:
\begin{enumerate}
    \item \textbf{Time-aware evaluation of agentic forecasters.} We introduce a temporal benchmark and with/without-search ablation that measures model performance at five points in each market's lifecycle rather than at a single snapshot.
    \item \textbf{Descriptive evidence of temporal and difficulty-dependent performance.} Models are most competitive early in a market's life and on high-uncertainty markets, but substantially less competitive late in the lifecycle and on strong-consensus markets.
    \item \textbf{Evidence for selective rather than uniform tool use.} Web search improves pooled BSS for all 10 models overall, but degrades performance in 12\% of model--checkpoint conditions and shows strong category-specific heterogeneity.
    \item \textbf{A simple ensemble baseline.} Model--market error correlations are modest ($r = 0.19$--$0.41$), and simple two-model averaging reduces loss relative to the best single model, though no pair surpasses the market overall.
\end{enumerate}

\section{Related Work}

\textbf{LLM forecasting and prediction benchmarks.} KalshiBench \citep{Nel2025} evaluates LLMs on thousands of Kalshi markets at fixed snapshots and highlights systematic miscalibration. ForecastBench \citep{KargerEtAl2024} and Prophet Arena \citep{YangEtAl2025b} move toward dynamic evaluation on live markets, while Mirai \citep{YeEtAl2024} emphasizes event-driven time sensitivity. TruthTensor \citep{ShahabiEtAl2026} evaluates LLMs under market drift with emphasis on human imitation, but does not control for tool access or analyze within-market lifecycle dynamics. Our work complements this literature with a controlled tool-use ablation across temporal checkpoints and difficulty regimes. Adjacent work studies forecasting-specific reliability via consistency checks \citep{PalekaEtAl2024} and uncertainty-aware training \citep{DamaniEtAl2025}.

\textbf{Agentic AI for financial decision-making.} Beyond forecasting, benchmarks increasingly measure end-to-end agent performance in market interaction, including multi-market trading \citep{QianEtAl2025,YuEtAl2025} and financial decision-making \citep{LiEtAl2025}. These stress execution and feedback loops; we focus on the upstream question of forecast reliability---whether an agent's probabilistic prediction should be trusted before any trading decision is made.

\textbf{Tool use and temporal drift.} Tool access can improve task performance \citep{SchickEtAl2023}, but tool-augmented workflows can distort confidence \citep{XuanEtAl2026} and deeper reasoning can degrade calibration \citep{LacombeEtAl2025}. Broader concerns about temporal drift in LLM behavior motivate evaluation protocols that are explicitly time-aware \citep{Khairnar2025,ZhangEtAl2025}. Our temporal checkpoint design directly tests these stability and reliability issues in a high-stakes forecasting setting.

\textbf{Prediction markets as baselines.} Prediction markets are generally well-calibrated across many domains \citep{WolfersZitzewitz2004,ArrowEtAl2008}, and crowd forecasting systems span markets, polls, and elite forecasters \citep{AtanasovEtAl2022}. We use contemporaneous market prices as the baseline, consistent with the view that market-implied probabilities are strong default forecasts \citep{WolfersZitzewitz2004,Fama1970}.

\section{Method}

\subsection{Dataset: CFTC-Regulated Binary Markets}

We collect 150 binary markets from Kalshi, a CFTC-regulated exchange where real capital is at risk. Each market resolves to YES/NO based on objective criteria (election results, economic releases, sports scores). We enforce \$5{,}000+ trading volume per market and require post-October 2025 resolution to ensure temporal separation from model training cutoffs. Concretely, we set the cutoff date to the latest publicly stated knowledge cutoff among the models we evaluate (October 2025), and keep only markets whose resolution occurs after that date to minimize outcome contamination (a KalshiBench-style filtering rule) \citep{Nel2025}. The dataset spans five categories (30 markets each): Politics, Sports, Macro-Economics, Science/Tech, and Financial markets. Table~\ref{tab:dataset} provides statistics.

\begin{table}[t]
\caption{TimeSeek dataset characteristics.}
\label{tab:dataset}
\centering
\small
\begin{tabular}{ll}
\toprule
Characteristic & Value \\
\midrule
Markets / Checkpoints / Predictions & 150 / 5 / 750 per condition \\
Categories & 5 (30 markets each) \\
Timeline & Oct 2025 -- Jan 2026 \\
Trading volume & \$5.6k -- \$41.3M (med.\ \$35k) \\
Duration (days) & 2 -- 337 (med.\ 89) \\
Outcomes & 73\% NO, 27\% YES \\
\bottomrule
\end{tabular}
\end{table}

\subsection{Temporal Sampling Protocol}

For each market, we define lifecycle progress
\[
p(t) = \frac{t - t_{\mathrm{open}}}{t_{\mathrm{close}} - t_{\mathrm{open}}}
\]
and sample at 5 checkpoints: Open+1 ($p = 0.01$), 25\%, 50\%, 75\%, and Close-1 ($p = 0.99$). For each sample, we record sample date, market price, and ground truth outcome, yielding 750 samples per model per condition (with/without search), totaling 15{,}000 predictions.

\subsection{Agentic Forecasting Architecture}

We implement agentic forecasters as a 4-node state machine: research (initialize context) $\rightarrow$ agent (decide to search or predict) $\rightarrow$ tools (execute searches) $\rightarrow$ forecast (extract structured prediction). The agent loops between agent and tools until it produces a final forecast.

When web search is enabled (agentic condition), agents access web search with date filtering (results restricted to before the sample date) to prevent future information leakage. When disabled (baseline condition), the model receives the identical prompt but cannot invoke any tools, isolating parametric knowledge. Market prices are withheld during prediction to prevent anchoring.

\subsection{Models Evaluated}

We evaluate 10 frontier LLMs: Claude Opus 4.5 \citep{Anthropic2025}, GPT-5.2 \citep{OpenAI2025}, Gemini 3 Pro \citep{Google2025}, Grok 4.1-fast \citep{xAI2025}, Kimi-k2 \citep{XuEtAl2025}, Kimi-k2.5 \citep{MoonshotAI2026}, DeepSeek v3.2 \citep{DeepSeekAI2025}, Intellect-3 \citep{PrimeIntellectTeam2025}, Trinity Large \citep{ArceeAI2026}, and Qwen3-235B \citep{YangEtAl2025a}. Each model predicts on 150 markets $\times$ 5 checkpoints $= 750$ samples per condition, yielding $750 \times 10$ models $\times 2$ conditions $= 15{,}000$ total predictions.

\subsection{Evaluation Metrics}

Brier Score (BS):
\[
\mathrm{BS} = \frac{1}{N}\sum_{i=1}^{N}(\hat{p}_i - y_i)^2,
\]
where $\hat{p}_i$ is predicted probability and $y_i \in \{0,1\}$ \citep{Brier1950}.

Brier Skill Score (BSS):
\[
\mathrm{BSS} = 1 - \frac{\mathrm{BS}_{\mathrm{model}}}{\mathrm{BS}_{\mathrm{market}}},
\]
using market price at prediction time as baseline \citep{Murphy1973}. Positive BSS indicates beating markets.

\section{Results}

Unless otherwise noted, the results below are descriptive point estimates without confidence intervals, so subgroup comparisons should be interpreted with appropriate caution.

\subsection{Temporal Degradation and Information Regimes}

Table~\ref{tab:bss-time} presents BSS for all models with web search. Four models outperform the market baseline at Open+1 (Claude +0.167, Kimi-k2.5 +0.118, GPT +0.085, Kimi-k2 +0.011), but all underperform it by Close-1 (BSS $< -0.7$). Performance declines monotonically across 9 of 10 models, with Kimi-k2 showing a slight recovery at Close-1.

\begin{table}[t]
\caption{BSS across temporal checkpoints with web search. Bold = beating market (BSS $> 0$).}
\label{tab:bss-time}
\centering
\footnotesize
\begin{tabular}{lrrrrrr}
\toprule
Model & Open+1 & 25\% & 50\% & 75\% & Close-1 & Pooled \\
\midrule
Claude Opus 4.5 & \pos{+0.167} & \pos{+0.059} & -0.136 & -0.165 & -0.829 & -0.068 \\
GPT-5.2 & \pos{+0.085} & -0.095 & -0.344 & -0.595 & -0.960 & -0.252 \\
Gemini 3 Pro & -0.084 & -0.250 & -0.373 & -0.510 & -0.734 & -0.312 \\
Grok 4.1-fast & -0.003 & -0.071 & -0.314 & -0.621 & -0.909 & -0.267 \\
Kimi-k2.5 & \pos{+0.118} & -0.022 & -0.311 & -0.579 & -0.942 & -0.214 \\
Kimi-k2 & \pos{+0.011} & -0.113 & -0.601 & -0.912 & -0.823 & -0.367 \\
DeepSeek v3.2 & -0.064 & -0.424 & -0.690 & -0.870 & -1.158 & -0.507 \\
Intellect-3 & -0.057 & -0.433 & -0.641 & -0.698 & -1.736 & -0.527 \\
Trinity Large & -0.128 & -0.275 & -0.581 & -0.684 & -1.716 & -0.496 \\
Qwen3-235B & -0.132 & -0.401 & -0.955 & -1.018 & -1.334 & -0.615 \\
\bottomrule
\end{tabular}
\end{table}

One plausible interpretation is that the benchmark spans two broad information regimes. Early in the market lifecycle (Open+1, 25\%), prices may reflect thinner participation, leaving more room for a web-enabled model to surface relevant public information that is not yet fully incorporated into price. Late in the lifecycle (75\%, Close-1), the market is a much stronger baseline after aggregating more trades and more diverse signals \citep{AtanasovEtAl2022}. Under this interpretation, the apparent crossover occurs between 25--50\% of the lifecycle for the strongest models (Claude, GPT) and earlier for weaker models. We treat this as a descriptive pattern rather than a causal identification of market microstructure.

\subsection{Web Search: A Double-Edged Sword}

Web search improves pooled BSS by 0.14--0.59 points across all models. However, computing $\Delta \mathrm{BSS} = \mathrm{BSS}_{\mathrm{agentic}} - \mathrm{BSS}_{\mathrm{baseline}}$ for every model-checkpoint pair (Table~\ref{tab:delta-time}) reveals that 6 of 50 conditions have negative $\Delta \mathrm{BSS}$. GPT-5.2 at Open+1, for example, drops from BSS +0.125 (baseline) to +0.085 (agentic), and Kimi-k2 is worse with search at three checkpoints (Open+1, 50\%, 75\%). These cases show that average gains from retrieval mask substantial heterogeneity across models and time.

\begin{table}[t]
\caption{$\Delta$BSS (agentic minus baseline). Bold = search helped. Shaded = search hurt---model was better without tools.}
\label{tab:delta-time}
\centering
\footnotesize
\begin{tabular}{lrrrrrr}
\toprule
Model & Open+1 & 25\% & 50\% & 75\% & Close-1 & Pooled \\
\midrule
Claude Opus 4.5 & \pos{+0.189} & \pos{+0.378} & \pos{+0.540} & \pos{+0.816} & \pos{+1.383} & \pos{+0.520} \\
GPT-5.2 & \negcell{-0.040} & \pos{+0.108} & \pos{+0.130} & \pos{+0.087} & \pos{+0.893} & \pos{+0.140} \\
Gemini 3 Pro & \negcell{-0.080} & \pos{+0.218} & \pos{+0.246} & \pos{+0.525} & \pos{+1.693} & \pos{+0.325} \\
Grok 4.1-fast & \pos{+0.214} & \pos{+0.457} & \pos{+0.721} & \pos{+0.679} & \pos{+1.710} & \pos{+0.592} \\
Kimi-k2 & \negcell{-0.025} & \pos{+0.234} & \negcell{-0.135} & \negcell{-0.146} & \pos{+1.388} & \pos{+0.135} \\
Kimi-k2.5 & \pos{+0.110} & \pos{+0.180} & \pos{+0.172} & \pos{+0.195} & \pos{+0.777} & \pos{+0.218} \\
DeepSeek v3.2 & \pos{+0.053} & \pos{+0.090} & \pos{+0.097} & \pos{+0.193} & \pos{+1.044} & \pos{+0.190} \\
Intellect-3 & \pos{+0.135} & \negcell{-0.049} & \pos{+0.107} & \pos{+0.662} & \pos{+0.729} & \pos{+0.228} \\
Trinity Large & \pos{+0.000} & \pos{+0.148} & \pos{+0.389} & \pos{+0.304} & \pos{+0.816} & \pos{+0.240} \\
Qwen3-235B & \pos{+0.037} & \pos{+0.280} & \pos{+0.063} & \pos{+0.101} & \pos{+1.416} & \pos{+0.245} \\
\bottomrule
\end{tabular}
\end{table}

This pattern is consistent with prior work suggesting that tool use can distort confidence and calibration in agentic workflows \citep{XuanEtAl2026}: more information processing is not uniformly beneficial. Further analysis shows a non-monotonic association between Claude's search count and BSS (Table~\ref{tab:claude-searches}), with the best observed bucket at 4--7 queries (+0.109). We interpret this pattern cautiously because query count is endogenous to the agent's behavior and may also reflect question difficulty.

\begin{table}[t]
\caption{Search count vs.\ BSS (Claude). Performance is non-monotonic across query-count buckets.}
\label{tab:claude-searches}
\centering
\small
\begin{tabular}{rrr}
\toprule
Searches & $n$ & BSS \\
\midrule
1--3 & 48 & -0.811 \\
4--7 & 152 & \pos{+0.109} \\
8--12 & 238 & +0.007 \\
13+ & 312 & -0.115 \\
\bottomrule
\end{tabular}
\end{table}

\textbf{Implication for selective tool use.} These results suggest that unrestricted tool access should not be assumed to help uniformly. A more realistic design target is an explicit tool-use policy that governs when to search, when to rely on parametric knowledge, and how many iterations to permit before forcing a prediction.

\subsection{LLMs Are Most Competitive on High-Uncertainty Markets}

Do LLMs and markets struggle on the same questions? We operationalize market difficulty using price distance from certainty: a market at 50\% represents maximum crowd uncertainty (toss-up), while one at 5\% or 95\% represents strong consensus (easy). Table~\ref{tab:difficulty} stratifies BSS by four such tiers. Because this definition is derived from market prices, we interpret the analysis as showing where models are relatively competitive against the market, not as a pure measure of intrinsic task difficulty.

Seven of ten models achieve positive BSS on toss-up markets. Claude leads at +0.301, followed by GPT (+0.217), Kimi-k2.5 (+0.179), and Kimi-k2 (+0.098). Three weaker models (Intellect-3, Trinity, Qwen3) remain negative even on toss-ups. The pattern reverses sharply for easy markets: all models show strongly negative BSS (from -0.692 to -1.680). Taken descriptively, this monotonic gradient suggests that current models are relatively most competitive when the market itself is most uncertain, and least competitive when market consensus is strong.

\begin{table}[t]
\caption{BSS stratified by market difficulty ($d = |p_m - 0.5|$): Easy $d>0.4$, Medium $0.2<d\leq0.4$, Hard $0.1<d\leq0.2$, Toss-up $d\leq0.1$. Bold = beating market.}
\label{tab:difficulty}
\centering
\footnotesize
\begin{tabular}{lrrrrr}
\toprule
Model & Easy & Medium & Hard & Toss-up & Overall \\
\midrule
Claude Opus 4.5 & -0.692 & -0.188 & \pos{+0.074} & \pos{+0.301} & -0.068 \\
GPT-5.2 & -1.069 & -0.433 & -0.001 & \pos{+0.217} & -0.252 \\
Gemini 3 Pro & -0.829 & -0.365 & -0.346 & \pos{+0.047} & -0.312 \\
Grok 4.1-fast & -0.905 & -0.256 & -0.201 & \pos{+0.002} & -0.267 \\
Kimi-k2.5 & -1.125 & -0.316 & \pos{+0.077} & \pos{+0.179} & -0.214 \\
Kimi-k2 & -0.906 & -0.555 & -0.296 & \pos{+0.098} & -0.366 \\
DeepSeek v3.2 & -1.308 & -0.669 & -0.413 & \pos{+0.041} & -0.507 \\
Intellect-3 & -1.680 & -0.625 & -0.108 & -0.109 & -0.527 \\
Trinity Large & -1.235 & -0.688 & -0.260 & -0.041 & -0.496 \\
Qwen3-235B & -1.502 & -0.857 & -0.238 & -0.123 & -0.615 \\
\midrule
\multicolumn{6}{l}{$n$ per tier: Easy: 254, Medium: 267, Hard: 100, Toss-up: 129} \\
\bottomrule
\end{tabular}
\end{table}

Pearson correlation between model absolute error and market absolute error (Table~\ref{tab:error-corr}) points in a similar direction: all models show low-to-moderate correlation ($r = 0.19$--$0.41$), suggesting only partial overlap between model and market error patterns. Intellect-3 shows the lowest correlation ($r = 0.193$) yet performs poorly on toss-ups, indicating that error diversity alone is not enough; accuracy matters too. Grok shows the highest correlation ($r = 0.408$), most closely tracking market pricing patterns.

\begin{table}[t]
\caption{Error correlation ($|\hat{p}-y|$ vs.\ $|p_m-y|$). Lower $r$ indicates less overlap between model and market error patterns. All $p < 10^{-6}$.}
\label{tab:error-corr}
\centering
\small
\begin{tabular}{lr}
\toprule
Model & Pearson $r$ \\
\midrule
Intellect-3 & 0.193 \\
Claude Opus 4.5 & 0.211 \\
Trinity Large & 0.218 \\
Kimi-k2.5 & 0.222 \\
DeepSeek v3.2 & 0.256 \\
Qwen3-235B & 0.261 \\
Kimi-k2 & 0.291 \\
GPT-5.2 & 0.336 \\
Gemini 3 Pro & 0.340 \\
Grok 4.1-fast & 0.408 \\
\bottomrule
\end{tabular}
\end{table}

\textbf{Contrarian disagreement by difficulty.} When Claude strongly disagrees with markets ($|\hat{p} - p_m| > 0.20$, $n = 303$), the outcome depends strongly on the market-uncertainty tier: contrarian cases on toss-ups achieve 80.9\% win rate (BSS +0.476, $n = 68$) and 65.4\% on hard markets (BSS +0.166, $n = 52$), but lose on easy markets (19.8\% win rate, BSS -0.720, $n = 81$). Within this benchmark, disagreement appears more informative when the market is uncertain than when consensus is strong, though we do not treat this offline analysis as a validated trading rule.

\textbf{Ensemble potential.} The low-to-moderate error correlations ($r = 0.19$--$0.41$) suggest some room for diversification via simple averaging \citep{BatesGranger1969}. The best two-model pair (Claude + Kimi-k2.5) reduces Claude's pooled BSS loss by 40\% (from -0.068 to -0.041), though no pair achieves positive overall BSS. All top ensembles include Claude, suggesting that base model quality still dominates any ensemble gains.

\subsection{Category-Level Heterogeneity in Search Impact}

Table~\ref{tab:category-bss} shows category-specific performance across all 10 models. Claude is strongest in Sports (+0.294) and Macro (+0.163); Kimi-k2.5 also shows positive Sports BSS (+0.124), and GPT and Gemini show positive Macro BSS. By contrast, all models are negative in Politics and Financial markets. A plausible interpretation is that these categories contain more information that is hard to recover reliably with periodic public-web retrieval alone.

\begin{table}[t]
\caption{BSS by category (agentic condition). Bold = beating markets.}
\label{tab:category-bss}
\centering
\footnotesize
\begin{tabular}{lrrrrr}
\toprule
Model & Politics & Sports & Macro & Sci/Tech & Financial \\
\midrule
Claude Opus 4.5 & -0.379 & \pos{+0.294} & \pos{+0.163} & -0.154 & -0.609 \\
GPT-5.2 & -0.484 & -0.088 & \pos{+0.021} & -0.378 & -0.482 \\
Gemini 3 Pro & -0.598 & -0.227 & \pos{+0.033} & -0.575 & -0.217 \\
Grok 4.1-fast & -0.427 & -0.197 & -0.182 & -0.190 & -0.426 \\
Kimi-k2.5 & -0.596 & \pos{+0.124} & -0.066 & -0.213 & -0.625 \\
Kimi-k2 & -0.421 & -0.342 & -0.023 & -0.369 & -0.788 \\
DeepSeek v3.2 & -0.703 & -0.110 & -0.333 & -0.596 & -1.182 \\
Intellect-3 & -0.706 & -0.255 & -0.275 & -0.569 & -1.131 \\
Trinity Large & -0.376 & -0.231 & -0.342 & -0.605 & -1.232 \\
Qwen3-235B & -1.020 & -0.046 & -0.601 & -0.719 & -1.158 \\
\bottomrule
\end{tabular}
\end{table}

Search impact also varies sharply by category. Table~\ref{tab:category-delta} presents $\Delta$BSS (agentic minus baseline) for all models. Search helps all 10 models in Macro (avg $\Delta$BSS +0.96) and Sports (avg +0.39), but is negative for most models in Politics (7 of 10; avg $\Delta$BSS -0.08) and Sci/Tech (6 of 10; avg -0.03). The most striking case is GPT-5.2 in Politics: its baseline nearly matches market accuracy (BSS -0.005 without search), but search degrades it to -0.484. Financial markets are evenly split (5 helped, 5 hurt), and Grok is the only model with positive $\Delta$BSS across all five categories. Overall, the evidence favors category-aware tool policies over a uniform ``always search'' rule.

\begin{table}[t]
\caption{$\Delta$BSS by category (agentic minus baseline). Bold = search helped. Shaded = search hurt. Bottom row: models helped / total and category average.}
\label{tab:category-delta}
\centering
\footnotesize
\begin{tabular}{lrrrrr}
\toprule
Model & Politics & Sports & Macro & Sci/Tech & Financial \\
\midrule
Claude Opus 4.5 & \negcell{-0.107} & \pos{+0.657} & \pos{+1.571} & \pos{+0.217} & \pos{+0.072} \\
GPT-5.2 & \negcell{-0.479} & \pos{+0.293} & \pos{+0.870} & \negcell{-0.170} & \pos{+0.074} \\
Gemini 3 Pro & \pos{+0.135} & \pos{+0.218} & \pos{+1.084} & \negcell{-0.267} & \pos{+0.604} \\
Grok 4.1-fast & \pos{+0.209} & \pos{+0.440} & \pos{+1.392} & \pos{+0.263} & \pos{+0.792} \\
Kimi-k2.5 & \negcell{-0.384} & \pos{+0.660} & \pos{+0.711} & \negcell{-0.113} & \negcell{-0.129} \\
Kimi-k2 & \negcell{-0.030} & \pos{+0.252} & \pos{+0.868} & \negcell{-0.169} & \negcell{-0.430} \\
DeepSeek v3.2 & \negcell{-0.240} & \pos{+0.301} & \pos{+0.845} & \pos{+0.051} & \negcell{-0.160} \\
Intellect-3 & \negcell{-0.018} & \pos{+0.230} & \pos{+0.588} & \pos{+0.053} & \pos{+0.302} \\
Trinity Large & \pos{+0.336} & \pos{+0.349} & \pos{+0.806} & \negcell{-0.179} & \negcell{-0.260} \\
Qwen3-235B & \negcell{-0.202} & \pos{+0.530} & \pos{+0.856} & \negcell{-0.024} & \negcell{-0.203} \\
\midrule
\multicolumn{1}{l}{\# helped / avg} & 3/10; -0.08 & 10/10; +0.39 & 10/10; +0.96 & 4/10; -0.03 & 5/10; +0.07 \\
\bottomrule
\end{tabular}
\end{table}

\section{Discussion}

\subsection{From ``forecasting'' to selective decision: when to defer to markets}

Our results suggest a useful framing is not ``LLMs vs.\ markets'' as a single contest, but a selective decision problem that chooses when to (i) forecast directly, (ii) search then forecast, or (iii) defer to the market price. This is closely related to selective prediction / classification-with-rejection, where a system trades coverage for lower risk and more reliable probabilistic outputs \citep{GeifmanElYaniv2017,FrancPrusa2023,CortesEtAl2016,CortesEtAl2024,MaoEtAl2024}. In our setting, deference corresponds to using the contemporaneous market-implied probability as a default baseline, consistent with the view that prediction markets often provide strong probabilistic forecasts in many domains \citep{WolfersZitzewitz2004,ArrowEtAl2008,PlottSunder1988}, while still acknowledging that prices only partially identify mean beliefs \citep{Manski2006}. This perspective turns TimeSeek into a problem of learning a gating function $g(x) \in \{\mathrm{defer}, \mathrm{predict}, \mathrm{search}\}$, potentially with a search budget, that optimizes a proper scoring objective \citep{GneitingRaftery2007}. It also suggests that coverage--risk tradeoffs may be more decision-relevant than unconditional BSS alone \citep{GeifmanElYaniv2017,FrancPrusa2023}. Post-hoc calibration methods such as temperature scaling and Platt scaling could map raw model scores into more reliable probabilities before a gating decision \citep{GuoEtAl2017,Platt1999,KuleshovEtAl2018}, and conformal methods offer another route to coverage guarantees under stronger assumptions \citep{ShaferVovk2008}.

\subsection{Why temporal crossover is expected: aggregation, frictions, and underreaction}

The late-stage degradation is consistent with classic theories of information aggregation and frictions, though our benchmark does not directly identify those mechanisms. In prediction markets, prices can aggregate dispersed information and heterogeneous beliefs \citep{WolfersZitzewitz2004,PlottSunder1988}, even if that aggregation is imperfect and prices only partially identify the full distribution of beliefs \citep{Manski2006}. Early in a contract's life, liquidity and participation can be thin, and prices may underreact to public signals due to limited attention, participation constraints, or capital frictions \citep{OttavianiSorensen2015,ShleiferVishny1997,DeLongEtAl1990}. These are precisely the settings in which web-enabled agents may add value by surfacing relevant public evidence that is available but not yet fully priced. As resolution approaches, however, markets typically incorporate real-time signals more quickly, and the marginal value of periodic retrieval shrinks relative to a continuously updated market. This framing also suggests one reason categories may differ: some domains may rely more heavily on information that is difficult to recover from public web search alone.

\subsection{Tool access is not monotone: retrieval can increase variance and miscalibration}

The finding that search sometimes hurts is consistent with emerging evidence that tool use can distort confidence and calibration in agentic workflows. Evidence tools such as web search can inject noisy, conflicting, or low-veracity signals that increase variance unless the agent has strong verification or calibration mechanisms \citep{XuanEtAl2026}. At the same time, \citet{LacombeEtAl2025} report calibration gains from search-augmented generation in general QA settings, which suggests that the value of retrieval is likely task- and domain-dependent rather than universal.

Our category-level analysis sharpens that point. Search helps uniformly in Macro and Sports, but is often neutral or harmful in Politics and Sci/Tech. We view this as consistent with a public-vs-private information story: some categories may be better matched to periodic public-web retrieval than others. The non-monotonic relationship between Claude's query count and performance points in the same general direction, but should be interpreted cautiously because search depth is itself a behavioral outcome. Taken together, these results motivate treating search as a budgeted, category-aware resource, potentially with explicit stopping rules, verification checks, and post-retrieval calibration \citep{PalekaEtAl2024,GneitingRaftery2007,GuoEtAl2017,DamaniEtAl2025}.

\subsection{Ensembling and market--model hybrids: why simple averages help}

That a two-model average reduces BSS loss is consistent with the forecasting-combination literature: combining partially independent predictors often reduces mean squared error, and simple averaging can be robust to weight misspecification \citep{BatesGranger1969,Clemen1989,Bunn1988,deMenezesEtAl2000}. In our setting, a natural next step is a conditional hybrid over the market and multiple models, where weights depend on temporal position and uncertainty tier. This suggests a concrete research direction: learn conditional weights, or a gating policy, that optimize proper scoring rules while controlling calibration and abstention behavior \citep{GneitingRaftery2007,GeifmanElYaniv2017,FrancPrusa2023}.

\subsection{Limitations}

\textbf{Statistical uncertainty and subgroup size.} We report point estimates throughout. With 150 markets, the benchmark supports strong directional patterns, but several subgroup analyses---especially category $\times$ time and difficulty slices---would benefit from bootstrap confidence intervals or other uncertainty estimates. Larger samples would also support more stable learned gating or weighting policies \citep{GeifmanElYaniv2017,deMenezesEtAl2000}.

\textbf{Market-defined difficulty and baseline heterogeneity.} Our difficulty tiers are derived from market prices, and BSS also uses the contemporaneous market as the baseline. The analysis is therefore best read as identifying where models are relatively competitive against the market, not as a clean decomposition of intrinsic task difficulty. More broadly, market prices embed heterogeneous belief aggregation and may underreact depending on liquidity, participation, and contract microstructure \citep{Manski2006,OttavianiSorensen2015,ShleiferVishny1997}.

\textbf{Offline evaluation rather than live trading.} We compare model forecasts to contemporaneous market prices, but we do not model execution, transaction costs, slippage, latency, or market impact. The results should therefore be interpreted as evidence about forecast quality relative to a market baseline, not as a validated trading strategy evaluation.

\textbf{Tooling scope and outcome skew.} We study only web search; richer tool suites may shift the observed crossover \citep{XuanEtAl2026}. The NO-heavy outcome distribution (73\%) can also interact with conservative prediction strategies and calibration \citep{GneitingRaftery2007,GuoEtAl2017}.

\subsection{Future Work}

\textbf{Learn the gate.} Learn a policy $g(x)$ that chooses defer/predict/search and sets a retrieval budget, optimizing proper scoring rules while controlling calibration and abstention \citep{GneitingRaftery2007,GeifmanElYaniv2017,FrancPrusa2023,XuanEtAl2026}.

\textbf{Streaming forecasters.} Replace one-shot retrieval with periodic updates and sequential decision-making, aligning agent updates with the market's continuous aggregation while explicitly managing variance and calibration drift \citep{OttavianiSorensen2015,XuanEtAl2026}.

\textbf{Mechanism-aware training sandboxes.} Many prediction markets rely on market-making mechanisms such as LMSR; integrating mechanism design and learning could enable controlled experiments on aggregation efficiency and agent impact \citep{Hanson2003,PlottSunder1988}.

\textbf{Market--model combination rules.} Extend from fixed averages to conditional combination schemes informed by forecast-combination theory and abstention, with guarantees on calibration/coverage \citep{BatesGranger1969,Clemen1989,deMenezesEtAl2000,ShaferVovk2008}.

\textbf{Post-training via market-resolved rewards.} Market outcomes provide verifiable binary labels at scale, making them natural reward signals for reinforcement learning. Fine-tuning forecasters with Brier-based RL rewards \citep{DamaniEtAl2025} on resolved markets could directly optimize calibration, with the gating policy learned jointly to control when the agent searches, predicts, or defers.

\section*{Reproducibility}

Code and data for TimeSeek are available in the accompanying repository at \url{https://github.com/coys17/timeseek-anonymous.git}. The repository includes the forecasting harness, temporal sampling pipeline, evaluation scripts, and the dataset of 150 Kalshi markets across five checkpoints.

\bibliographystyle{iclr2026_conference}
\bibliography{iclr2026_conference}

\end{document}